\documentclass{article}

\usepackage[main, final]{neurips_2026}
\usepackage{graphicx}

\usepackage[utf8]{inputenc} 
\usepackage[T1]{fontenc}    
\usepackage{hyperref}       
\usepackage{url}            
\usepackage{booktabs}       
\usepackage{amsfonts}       
\usepackage{nicefrac}       
\usepackage{microtype}      
\usepackage{xcolor}         

\usepackage{microtype}
\usepackage{graphicx}
\usepackage{subcaption}
\usepackage{booktabs} 
\usepackage{multirow}
\usepackage{amsmath}
\setcitestyle{square,numbers}

\title{Towards Effective and Transferable Physical Camouflage against Multi-View BEV-based 3D Perception in Autonomous Driving}

\author{
  Linye Lyu, Jiawei Zhou, Daojing He, Yu Li \\
  Harbin Institute of Technology, Shenzhen\\
  Zhejiang University\\
  \texttt{\{lyulinye, zhoujiawei6666, yu.li.sallylee\}@gmail.com} \\
  \texttt{hedaojinghit@163.com}\\
}

\begin{document}

\maketitle

\begin{abstract}
Modern autonomous driving systems widely adopt multi-view Bird’s-Eye-View (BEV) based 3D perception models due to their superior performance. Despite their success, the robustness of these models against adversarial camouflage attacks remains largely unexplored. Existing camouflage attacks primarily target single-view 2D detectors, limiting their effectiveness in multi-view 3D perception settings. To address this gap, we propose BEVCA, a novel adversarial camouflage generation framework tailored for multi-view BEV-based 3D perception. Our framework integrates a BEV-feature-based adversarial loss with a multi-view neural rendering module, enabling effective and transferable camouflage attacks across different models and tasks. Extensive experiments in both digital and physical settings demonstrate that BEVCA outperforms state-of-the-art baselines and exhibits strong robustness under real-world conditions. Our code is available at \href{https://github.com/WAVE-AT-IDEA/BEVCA}{https://github.com/WAVE-AT-IDEA/BEVCA}. 

\end{abstract}

\section{Introduction}

Vision-based autonomous driving (AD) systems have been widely adopted by major vehicle manufacturers, such as Tesla \cite{teslaFSD}, due to their competitive performance. These systems utilize multiple camera sensors surrounding the ego vehicle to perceive the nearby environment with deep learning models \cite{philion2020lift, zhou2022cross, xu2022cobevt, huang2021bevdet,wang2022detr3d, li2022BEVFormer}. However, despite the significant progress, current research shows that deep neural networks (DNNs) are vulnerable to adversarial examples \cite{szegedy2013intriguing}, which threaten the safety-critical nature of AD systems.

\begin{figure}[t]
\centering
\includegraphics[width=\columnwidth]{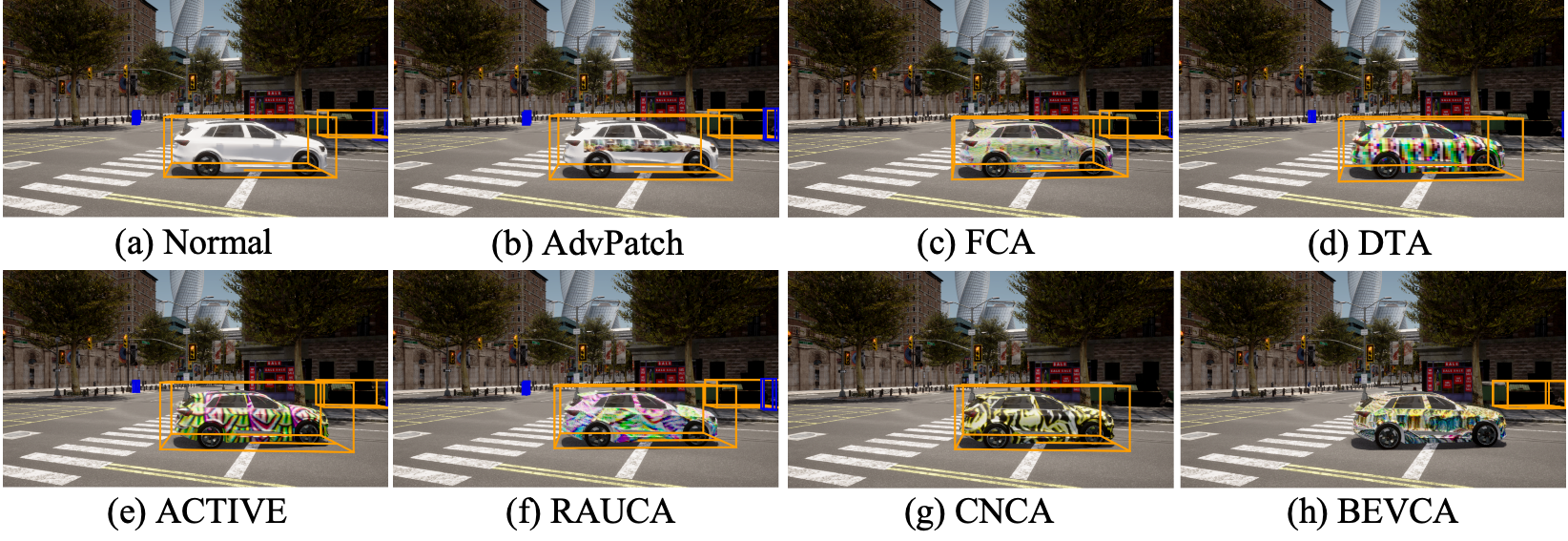}
\caption{Comparison of  3D detection results using different textures on the photo-realistic simulation environment, where only the adversarial camouflage generated by our BEVCA framework successfully evades detection. (a) A vehicle with normal painting; (b) AdvPatch \cite{wang2025unified}. (c)-(g) are the top-performed methods against 2D detection: FCA \cite{wang2022fca}, DTA \cite{Suryanto_2022_CVPR}, ACTIVE \cite{Suryanto_2023_ICCV}, RAUCA \cite{zhou2024rauca}, and CNCA \cite{lyu2024cnca} respectively; (h) Our method BEVCA.}
\label{fig:intro}
\end{figure}
Prior attacks mainly focus on the single-view 2D vehicle detection task. These methods \cite{wang2022fca, Suryanto_2022_CVPR, Suryanto_2023_ICCV, zhou2024rauca, lyu2024cnca,zhou2024toward} use the Carla simulator \cite{dosovitskiy2017carla} to generate vehicle images with various transformations to enhance the attack robustness. Besides, they leverage a differentiable neural renderer \cite{kato2018neural, ravi2020pytorch3d} to enable direct adversarial camouflage optimization against the target model.

Despite impressive progress in 2D attacks, current state-of-the-art vehicle detection methods in autonomous driving are dominated by multi-view 3D perception models. As a result, the current single-view 2D attack pipelines cannot be directly applied to the multi-view 3D perception models. Some works, such as \cite{zhu2023understanding, wang2025unified}, explore attacking multi-view 3D perception models with adversarial patches by minimizing detection scores. However, the adversarial patches are inherently not robust against various viewing angles since they cannot fully cover the target vehicle, leading to suboptimal attack performance. Besides, attacks in the real world are mostly black-box settings. Yet, all previous methods rely on the target model's detection scores to optimize the adversarial texture, thereby limiting their attack transferability.

To tackle the above challenges, we propose BEVCA, a novel framework for generating adversarial camouflage against multi-view BEV-based 3D perception models. Our key observation is that most state-of-the-art multi-view 3D perception models \cite{philion2020lift, huang2021bevdet, wang2022detr3d, li2022BEVFormer} rely on Bird's-Eye-View (BEV) features, a unified 2D spatial representation that supports various downstream tasks in AD systems \cite{hu2023planning, zheng2024genad}. This insight suggests that attacking BEV features could yield transferable adversarial effects across different models and tasks. However, we find that directly optimizing camouflage using naive BEV features is ineffective (see Section~\ref{ab-attack-table3}). The main challenge is to identify the BEV feature regions that are highly sensitive to the target vehicle’s texture . To address this, we propose a dedicated masking technique that accurately locates the BEV feature areas highly relevant to the target car and design adversarial loss accordingly. In addition, we extend the current single-view neural rendering to multi-view neural rendering to generate realistic, consistent adversarial vehicle images across multiple camera views, enabling gradient-based texture optimization. The combination of the BEV-feature-based attack and multi-view neural rendering enables highly effective, transferable adversarial camouflage across diverse BEV-based perception tasks and models.

Our main contributions are summarized as follows:
\begin{itemize}
    \item We present a BEV-feature-based Camouflage Attack (BEVCA), a framework for generating adversarial camouflage against multi-view BEV-based 3D perception models. It achieves effective and transferable adversarial camouflage attacks using BEV-feature-level attacks and multi-view neural rendering.
    \item We propose a novel mask-enhanced BEV-based adversarial loss that accurately targets car-texture-sensitive regions of the BEV, enabling strong transferability across diverse BEV-based 3D perception models and downstream tasks.
    \item We extend single-view neural rendering to multi-view neural rendering, producing realistic, consistent vehicle images accross multiple viewpoints relative to the ego vehicle, facilitating efficient gradient-based adversarial camouflage optimization.
\end{itemize}

We conduct a comprehensive evaluation with various multi-view BEV-based 3D perception models. The results show that our adversarial camouflage outperforms existing baselines in black-box settings, achieving average improvements of 36.2\% and 21.6\% in 3D object detection and segmentation, respectively. Furthermore, we also validate the robustness of our approach in the physical world, achieving an average attack success rate of 51.18\% in black-box settings.

\section{Related work and threat model}

\subsection{Related work}

In this section, we first introduce multi-view 3D perception, then review adversarial camouflage attacks on vehicle detection, and finally define the threat model for our scenario. 

\textbf{Multi-view 3D Perception.} The development of large-scale multi-view datasets such as NuScenes \cite{Caesar2020NuScens} and Waymo \cite{Sun2020Waymo} has significantly accelerated progress in multi-view 3D perception. Bird’s-Eye-View (BEV) models \cite{philion2020lift, huang2021bevdet, wang2022detr3d, li2022BEVFormer} have demonstrated state-of-the-art performance across various autonomous driving (AD) tasks. These models extract and aggregate information from multi-view driving images to construct a unified BEV feature representation of the environment, which can support various downstream tasks, including 3D object detection, map segmentation, and even end-to-end driving systems \cite{hu2023planning, zheng2024genad}.

\textbf{Adversarial Camouflage} Accurate vehicle detection is critical for the safety of AD systems, motivating a growing interest in studying the robustness of vehicle detectors with adversarial attacks. These attacks can be categorized as single-view 2D attacks and multi-view 3D attacks. Existing 2D attack methods leverage 3D simulators such as Carla \cite{dosovitskiy2017carla} to generate target vehicle images under various transformations for optimizing adversarial textures. Then, these methods \cite{wang2022fca, Suryanto_2022_CVPR, Suryanto_2023_ICCV, zhou2024rauca, lyu2024cnca} employ the differentiable neural renderer \cite{kato2018neural, ravi2020pytorch3d} to directly optimize adversarial textures via gradient backpropagation. Besides,  \cite{Suryanto_2022_CVPR, Suryanto_2023_ICCV, zhou2024rauca, lyu2024cnca} further incorporate realistic environmental features, such as shadows and fog, to mitigate the domain gap between simulation and real-world conditions, achieving better attack performance. Regarding multi-view 3D attacks, previous works attempt to generate the adversarial patches on the target vehicle.  \cite{zhu2023understanding, wang2025unified} utilize the 3D annotations of the target vehicle to transform the 2D adversarial patches so that they look realistic and consistent in the multi-view image setting, while \cite{li2024adv3d} utilizes NeRF to generate adversarial texture on the vehicle. 

In contrast to prior works, our framework introduces a novel combination of multi-view neural rendering and BEV-feature-based adversarial attack to enable effective and transferable attacks on multi-view BEV-based 3D perception models.

\subsection{Threat Model} 

\textbf{Targeted Model:} We consider attackers targeting multi-view perception models that construct a BEV feature representation for downstream tasks like object detection and segmentation. These BEV perception models, for instance, BEVFormer \cite{li2022BEVFormer}, are widely adopted in the current state-of-the-art multi-view 3D perception tasks in autonomous driving \cite{hu2023planning,zheng2024genad}. 

\textbf{Attacker:} The attacker's goal is to reduce the perception performance of the target models towards the camouflaged vehicle, subject to physical constraints and consistently effective across various viewpoints. This is achieved by altering the vehicle's appearance with adversarial camouflage textures. The attack operates in two distinct phases: firstly, \textit{{\textbf{offline optimization}}} where attacker leverages the simulation environment (e.g., CARLA) to obtain ground-truth target vehicle positions and generate perfectly paired no-vehicle reference scenes, and performs gradient-based optimization against the victim model (white-box). This information is strictly confined to the offline preparation phase. Secondly, \textit{{\textbf{passive deployment}}}, where the adversarial texture is printed and physically applied to the target vehicle. In the black-box setting, the attacker additionally assumes only that the victim model relies on BEV feature representation, with no internal access to model weights or intermediate features. In both settings, the attacker cannot modify training data, model weights, or sensor configurations.

\section{Methods}
In this section, we first present an overview of our framework for generating effective and transferable adversarial camouflage against multi-view BEV-based 3D perception models. Then, we describe the details of the essential components. 

\begin{figure*}[t]
\centering
\includegraphics[width=\textwidth]{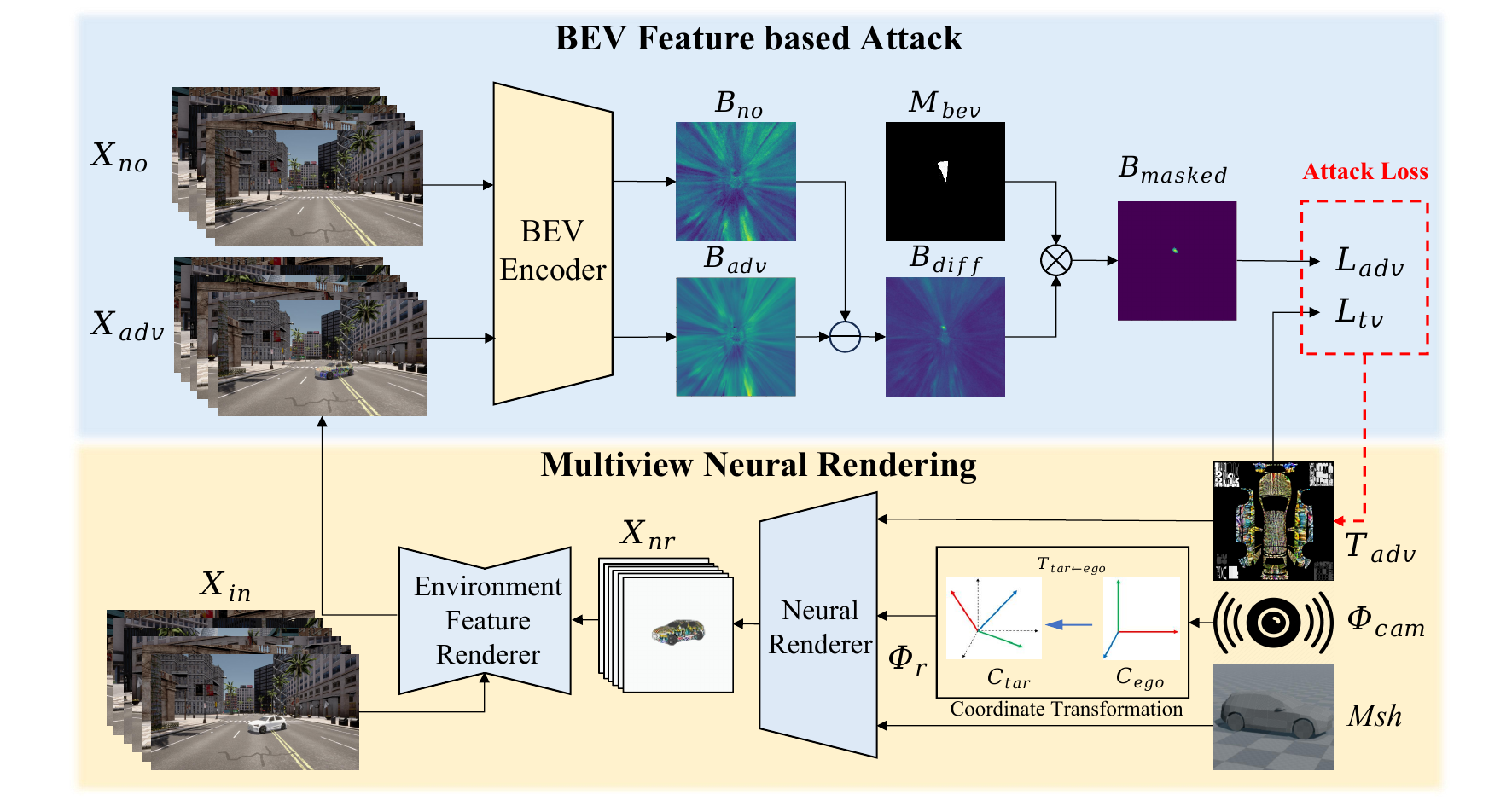}
\caption{The overview of BEVCA. Our framework consists of multi-view neural rendering and BEV-feature-based attack modules. The multi-view neural rendering outputs the multi-view camouflage vehicle images from the ego vehicle perspective. The BEV-feature-based attack locates the BEV feature regions that are highly associated with the target vehicle with a designed mask. The combination of both modules enables effective and transferable camouflage attacks against multi-view 3D perception models. 
  }
\label{pipeline}
\end{figure*}

\subsection{Overview}

Figure~\ref{pipeline} illustrates the overall framework for generating adversarial camouflage. We first construct a multi-view image dataset of a target vehicle captured from the ego vehicle's perspective using the Carla simulator \cite{dosovitskiy2017carla}, following the NuScenes \cite{Caesar2020NuScens} dataset format.  This dataset contains the multi-view images \(X_{in}\), camera configurations  \(\Phi_{cam}\), and the transformation matrices \(T_{tar \leftarrow ego}\) that map ego coordinates to target vehicle coordinates. Our multi-view neural renderer first uses a neural renderer \(NR\), which takes the adversarial texture \(T_{adv}\), 3D vehicle mesh  \(Msh\), camera configurations \(\Phi_{cam}\), and the transformation matrices \(T_{tar \leftarrow ego}\) as input, and produces the multi-view rendered adversarial target vehicle images \(X_{nr}\).
Subsequently, both \(X_{nr}\) and \(X_{in}\) are input into an Environmental Feature Renderer (EFR). We use the pretrained EFR from \cite{lyu2024cnca} to extract environmental features from \(X_{in}\) and render them together with the background to output the realistic multi-view camouflaged target vehicle images \(X_{adv}\). To summarize:
\begin{gather}
X_{adv}=EFR(X_{in}, NR(Msh, T_{adv}, \Phi_{cam}, T_{tar \leftarrow ego} ))\label{1}
\end{gather}

Instead of attacking task-specific output scores (e.g., detection confidence), we target the BEV feature representation to improve attack transferability. A key challenge is identifying the regions within the BEV feature map that correspond to the target vehicle’s presence in the multi-view images. To this end, we construct a \textit{no-vehicle} multi-view dataset \(X_{no}\) under the identical settings as \(X_{in}\), except that the target vehicle is removed. We feed \(X_{adv}\) and \(X_{no}\) into a BEV encoder to obtain the corresponding BEV features:
\begin{gather}
B_{adv}= BEV(X_{adv})\label{2}\\
B_{no}= BEV(X_{no})\label{3}
\end{gather}

Since the only difference between \(X_{adv}\) and \(X_{no}\) is the presence of the target vehicle, we hypothesize that the difference between \(B_{adv}\) and \(B_{no}\) reflects the target vehicle's impact on BEV feature space:
\begin{gather}
B_{diff}= |B_{adv} - B_{no}|\label{4}
\end{gather}

Empirically, we observe that the prominent regions in \(B_{adv}\) are geometrically aligned with the target vehicle’s ground-truth location. Based on this observation, we apply a BEV mask \(M_{bev}\) to \(B_{diff}\) , locating the affected regions \(B_{masked}\) :
\begin{gather}
B_{masked}= B_{diff} \cdot M_{bev}\label{5}
\end{gather}

We define an adversarial loss \(L_{adv}\)  based on \(B_{masked}\) to attack the BEV representation towards the target vehicle. Additionally, following \cite{sharif}, we employ a total variation loss \(L_{tv}\) on the adversarial texture image \(T_{adv}\) to ensure the naturalness of texture to human vision. The details of \(L_{adv}\) and \(L_{tv}\) will be discussed in the following section. Finally, our total loss function to optimize the adversarial texture is:
\begin{gather}
L_{total }=\alpha L_{adv}+\beta L_{tv},\label{6}
\end{gather}

where $\alpha$\label{alpha}, $\beta$\label{beta}  are the hyperparameters to control the contribution of each loss function.

\subsection{Multi-view Neural Rendering}

Previous works \cite{Suryanto_2022_CVPR, Suryanto_2023_ICCV, zhou2024rauca, lyu2024cnca} have demonstrated the effectiveness of adversarial camouflage against 2D object detectors by leveraging realistic neural rendering of textured vehicles. In the single-view setting, the adversarial vehicle is placed at the center of the image, and the rendering can be achieved using the object-centric rendering methods from \cite{ravi2020pytorch3d, kato2018neural}. These methods only require the relative camera pose with respect to the adversarial vehicle, which can be easily obtained during the camera initialization in CARLA. However, this pipeline is not directly applicable to multi-view 3D perception tasks. In such scenarios, multiple cameras are attached to the ego vehicle to observe the target vehicle. As a result, the camera pose relative to the target vehicle—required by the renderer—can no longer be retrieved through the simple initialization procedure used in the single-view pipeline. To support multi-view rendering, it is necessary to redesign the pipeline by computing the relative pose between each ego-mounted camera and the target vehicle, based on the geometric relationships among the ego vehicle, the target vehicle, and the cameras.

To address this, we extend the popular differentiable rendering library PyTorch3D \cite{ravi2020pytorch3d} as the naive renderer to support multi-view rendering. The naive renderer requires three key inputs: the 3D mesh of the vehicle, the adversarial texture, and the camera settings relative to the target vehicle. The key challenge is to compute the transformations \(T_{tar \leftarrow ego}\), which convert the camera settings of each camera coordinate system \(\Phi_{cam}\) to the target vehicle’s coordinate system:
\begin{equation}
\Phi_{r} = T_{tar \leftarrow ego} \cdot \Phi_{cam}
\end{equation}
where \(\Phi_{r}\) can be directly input to the naive neural renderer. These transformations ensure that each rendered view accurately reflects the camera's perspective towards the target vehicle. We apply these camera parameter transformations to all six ego-mounted cameras and perform image rendering as defined in Eq ~\ref{1} to obtain multi-view rendering images. The full mathematical details and transformation steps are provided in the appendix \ref{appendix:multi_view_neural_rendering} for interested readers.

\subsection{BEV-Feature-based Adversarial Attack}

\label{bev-feature-attack}

\begin{figure}[t]
\centering
\includegraphics[width=\columnwidth]{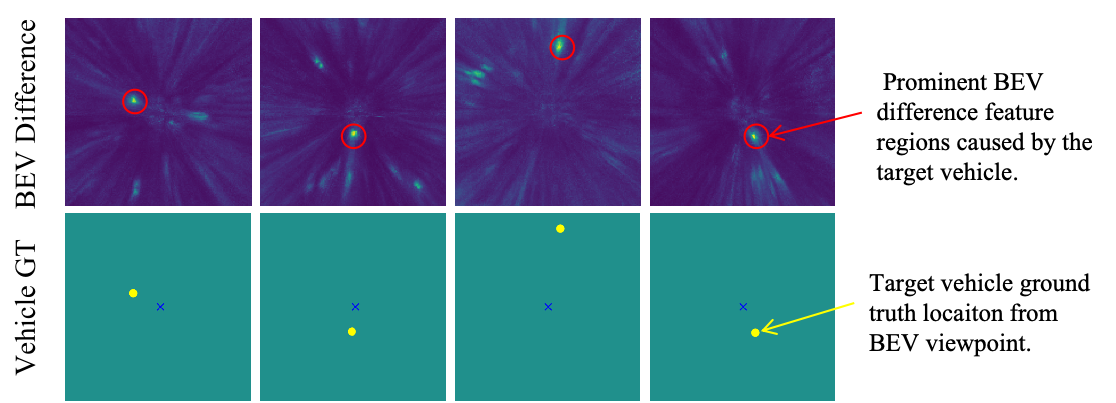}
\caption{Examples of BEV difference feature and corresponding vehicle ground truth images.  We can observe a strong correlation between BEV regions relevant to the target vehicle (marked by the red circles) and vehicle ground truth position (marked by the yellow points). }
\label{fig:bev_diff_vs_gt}
\end{figure}

The current state-of-the-art multi-view 3D perception models \cite{philion2020lift, huang2021bevdet, wang2022detr3d, li2022BEVFormer} rely heavily on BEV features to support downstream perception tasks in autonomous driving systems \cite{hu2023planning, zheng2024genad}. This motivated us to attack from the BEV feature to achieve an effective and transferable attack. However, since the BEV feature is high-dimensional, it is challenging to directly isolate the feature areas that are influenced by the target vehicle. To tackle this, we utilize the BEV feature extracted from the dataset with the same camera configurations but excluding the target vehicle. We analyze the difference between these two BEV features to locate the target vehicle's associated feature regions, which we used to calculate the adversarial loss.  

Specifically, as shown in Eq. \ref{2} and \ref{3}, we first obtain the BEV features \(B_{adv}\) and \(B_{no}\) from \(X_{adv}\) and \(X_{no}\) respectively. We compute their difference to obtain the \(B_{diff}\), which has the shape of \((H,W,C)\), where \(H\) and \(W\) define the BEV perception grid, and  \(C\) is the feature dimension.  To localize the affected regions, we average across the feature dimension C and visualize the resulting 2D heatmap. Empirically, as shown in Figure \ref{fig:bev_diff_vs_gt}, we observe that high-activation areas in this map tend to coincide with the ground-truth location of the target vehicle, suggesting a strong correlation.

Based on the above observation, we propose a BEV mask  \(\mathcal{M}_{bev}\) to effectively crop out the affected regions. We
leverage the annotated position of the target vehicle to build this mask. For convenience, we 
represent the target’s position in polar coordinates \((r_{tar}, \theta_{tar})\) related to the ego vehicle and define the mask region as:
\begin{equation}
\mathcal{M}_{bev} = \left\{ (r, \theta) \,\middle|\, |r - r_{\text{tar}}| \leq \Delta r,\ |\theta - \theta_{\text{tar}}| \leq \Delta \theta \right\}
\label{eq:13}
\end{equation}
where \(\Delta r\) and \(\Delta \theta\) are the tunable thresholds to control the mask area. In the ablation section, we evaluate the effectiveness of this mask strategy. 
 
 Once the masked BEV difference \(B_{masked}\) is obtained  (as shown in Eq. \ref{5}.  ), we define the adversarial  \( L_{adv} \)  as the mean of the non-zero elements in \( B_{masked} \):
\begin{equation}
L_{adv} = \frac{\sum\limits_{i} B_{masked, i}}{\sum\limits_{i} \mathbb{1}(B_{masked, i} \neq 0)}
\end{equation}

where \( \mathbb{1}(\cdot) \) is the indicator function, which returns 1 if the condition inside is true and 0 otherwise. The denominator represents the number of non-zero elements in \( B_{masked} \), and the numerator is the sum of all elements in \( B_{masked} \). 

Additionally, we also apply the Total Variation (TV) loss followed by \cite{sharif} to encourage naturalness of the texture towards human vision:
\begin{equation}
L_{tv} = \sum_{i, j} \left( x_{i, j} - x_{i+1, j} \right)^2 + \left( x_{i, j} - x_{i, j+1} \right)^2
\end{equation}

\section{Experiments}

\subsection{Experimental Settings}

\label{sec:experiment_setting}

\textbf{Datasets.} \label{dataset}We utilize the Carla simulator  \cite{dosovitskiy2017carla}  to generate datasets for our experiments. To have a comparative analysis with prior studies \cite{wang2022fca, Suryanto_2023_ICCV}, we select the Audi E-Tron as the target vehicle model.  We generate a dataset of 2500  samples for camouflage generation and a test dataset of 500 samples for camouflage evaluation. These datasets contain multi-view images of the target vehicle with various distances and view angles relative to the ego vehicle, across various background scenes. During our experiments, we found that the performance of the original targeted models degraded significantly in the simulation dataset compared to their reported results, as shown in Table~\ref {sim-attack-original-table}. This is because these models are trained on realistic datasets like nuScenes \cite{Caesar2020NuScens}. To alleviate the domain gap between simulation and realism, we generate a fine-tuning dataset of 500 samples with completely different settings from camouflage generation and test datasets to adapt these models to the digital domain. For the real-world evaluation, we print our camouflage in a 2D UV map image and stick it to a full-size Audi vehicle. We build a simple multi-view camera data collection platform to collect real-life multi-view data. We collect three physical datasets: one of the car with normal painting,  one of the car with CNCA texture, and one of the car with our camouflage. Each dataset has 150 samples with different azimuths and distances towards the ego position.


\textbf{Baselines.}
We compare BEVCA with five adversarial camouflage methods targeting 2D vehicle detection---FCA \cite{wang2022fca}, DTA \cite{Suryanto_2022_CVPR}, ACTIVE \cite{Suryanto_2023_ICCV}, RAUCA \cite{zhou2024rauca}, and CNCA \cite{lyu2024cnca}---and AdvPatch \cite{wang2025unified}, an adversarial patch method targeting 3D detection. For each baseline, we apply its officially released adversarial texture to the Audi E-Tron model and evaluate it using the same multi-view BEV perception pipeline. These baseline textures are not re-optimized or adapted to our experimental setting. 

\textbf{Evaluation metrics.}\label{evaluation-metrics} In digital experiment, to evaluate the effectiveness of the adversarial camouflage, we utilize the Average Precision (AP) score defined by the 2D center distance between the prediction and ground truth on the ground plane \cite{Caesar2020NuScens,li2022BEVFormer}  for 3D detection task and Intersection-over-Union (IoU) \cite{zhou2022cross} of vehicle class for BEV segmentation task, respectively.  In the real-world experiments, we report Attack Success Rate (ASR), defined as the percentage of target vehicles detected in the clean car painting that are no longer detected in the adversarial camouflage. Detection is determined using an IoU threshold of 0.5, with the same evaluation criteria applied to clean and adversarial cases.

\textbf{Target models.}\label{target-models} We choose the BEVFormer-base \cite{li2022BEVFormer} model as the white-box target model for adversarial camouflage generation. To evaluate the effectiveness and transferability of the optimized camouflage, we utilize a collection of 3D object detection and BEV segmentation models treated as black-box models. 3D object detection models include BEVFormer series (small, tiny, base-seg-det, small-seg-det) \cite{li2022BEVFormer} with different model configurations. The BEV segmentation models include Lift-Splat-Shoot \cite{philion2020lift}, Cross View Transformer \cite{zhou2022cross} and SinBEVT \cite{xu2022cobevt}. We train the original version of these models on the fine-tuning dataset with one epoch before evaluating on the test dataset.

\textbf{Optimization details.}\label{optim-details} We utilize the Adam optimizer with a learning rate of 0.01 for adversarial camouflage generation. The camouflage texture is initialized randomly and trained with 10 epochs. For the hyperparameters during camouflage optimization, we set the values of $\alpha$ and $\beta$ (see Eq. \ref{6})  to 1 and 1000 respectively, as shown in Figure \ref{fig:hp_beta_tuning}. For the BEV mask, we set the distance and angle thresholds (\(\Delta r, \Delta \theta\))  to 15 meters and 15 degrees (see Eq. \ref{eq:13}) after hyperparameter tunings. We conduct experiments on one NVIDIA RTX A800 80GB GPU with around 20 hours of camouflage optimization time.

\subsection{Attack Performance Evaluation}

\textbf{Digital attack on 3D perception models.} We compare our proposed method, BEVCA, against several state-of-the-art adversarial camouflage approaches, including FCA, DTA, ACTIVE, RAUCA, CNCA, and AdvPatch. To thoroughly evaluate the effectiveness and transferability of our method, we conduct extensive experiments across a variety of multi-view 3D perception models in both 3D object detection and BEV segmentation tasks. The results are shown in Table \ref{sim-attack-table1}, where BEVCA consistently outperforms all baselines. AdvPatch shows performance comparable to that of a normally painted vehicle, primarily due to its reliance on small, localized patches. FCA provides only marginal improvement, as it lacks the ability to model complex environmental interactions. In contrast, DTA, ACTIVE, RAUCA, and CNCA offer stronger performance, yet remain suboptimal because they are not explicitly designed for multi-view perception models. Our method surpasses all the baselines, achieving an average performance gain of approximately 36.2\% in black-box settings against the best previous baseline, demonstrating strong effectiveness and transferability.

We also provide some adversarial camouflage vehicle examples against 3D detection in different scenarios in the appendix section . These examples demonstrate that the vehicle with normal painting is accurately detected with well-aligned 3D bounding boxes. However, with our adversarial camouflage texture, the target vehicle becomes significantly harder to detect, often resulting in inaccurate bounding boxes or complete disappearance from the detection output.

For the BEV segmentation task, although the adversarial camouflage generated by BEVCA is specifically optimized for the BEVFormer base model, it consistently achieves the best attack performance across all segmentation models when compared to existing baselines. Notably, BEVCA attains an average improvement of approximately 21.6\% over the strongest previous baseline performance in black-box settings. These results further underscore the task transferability and robust generalization of our approach across diverse BEV-based perception tasks.

\begin{table*}[t]
\caption{Comparison of the effectiveness of attacks across various 3D detection and BEV segmentation models. The metrics for 3D detection and BEV segmentation are AP and IoU of the vehicle class, respectively.}
\label{sim-attack-table1}

\begin{center}
\begin{small}
\begin{sc}
\resizebox{1.0\textwidth}{!}{ 
\begin{tabular}{lcccccccc}
\toprule
\multirow{2}{*}{\textbf{Methods}} & \multicolumn{5}{c}{3D Detection} & \multicolumn{3}{c}{BEV Segmentation } \\
\cmidrule(lr){2-6} \cmidrule(lr){7-9}
                         & Base & Small & Tiny & Base-Seg-Det & Small-Seg-Det & LSS & CVT & SinBEVT \\
\midrule
Normal                   & 0.618 & 0.721 & 0.514 & 0.750 & 0.634 & 0.509& 0.609& 0.570\\
FCA                      & 0.593 & 0.696 & 0.512 & 0.721 & 0.595 & 0.442& 0.505&0.489\\
DTA                      & 0.578 & 0.663 & 0.469 & 0.684 & 0.591 & 0.372& 0.480&0.449\\
ACTIVE                   & 0.559 & 0.664 & 0.433 & 0.682 & 0.588 & 0.367& 0.481&0.436\\
RAUCA                    & 0.562 & 0.682 & 0.479 & 0.689 & 0.576 & 0.380& 0.459&0.427\\
CNCA                     & 0.567 & 0.626 & 0.404 & 0.634 & 0.523 & 0.306& 0.401&0.360\\
AdvPatch                 & 0.601 & 0.702 & 0.516 & 0.703 & 0.608 & 0.479& 0.579&0.533\\
\midrule
BEVCA                    & \textbf{0.191} & \textbf{0.285} & \textbf{0.381} & \textbf{0.329} & \textbf{0.331} & \textbf{0.253}& \textbf{0.307}& \textbf{0.273}\\
\bottomrule
\end{tabular}
}
\end{sc}
\end{small}
\end{center}

\end{table*}

\begin{table*}[t]
\caption{Attack success rates (ASR \%) by methods and models in the physical world evaluation.}
\label{phy-table1}
\begin{center}
\begin{small}
\begin{sc}
\begin{tabular}{lcccc}
\toprule
\textbf{Methods} & BEVFormer base & BEVFormer small & BEVDet-r50 & BEVDet-r50-cbgs \\
\midrule
CNCA & 30.00\% & 29.49\% & 42.80\% & 34.99\% \\
BEVCA  & 49.60\% & 46.99\% & 64.41\% & 42.13\% \\
\bottomrule
\end{tabular}
\end{sc}
\end{small}
\end{center}
\end{table*}





\begin{figure*}[htbp]
\centering
\includegraphics[width=1.0\textwidth]{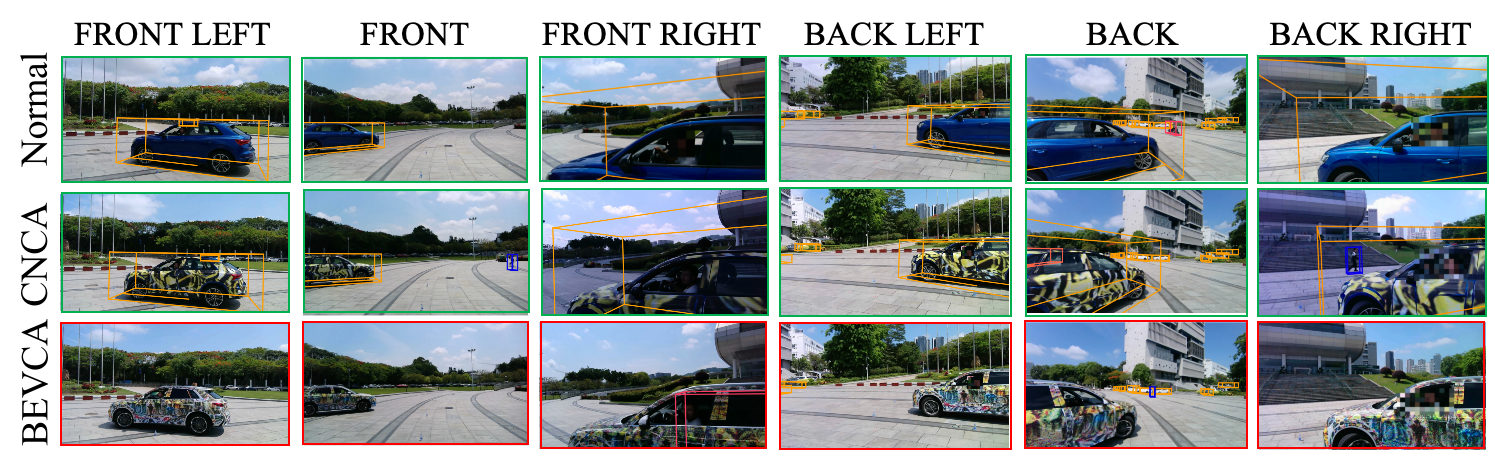}

\caption{Visualization of our camouflage successful attack examples against 3D object detection in the physical world from different camera viewpoints.}
\label{phy_exp_examples}
\end{figure*}

\textbf{Physical world evaluation.} We conduct the physical world evaluation by printing the camouflage in a 2D UV map format and sticking it to a full-size Audi car. To collect real multi-view data, we built a multi-view camera data collection platform following the camera setup from the nuScenes \cite{Caesar2020NuScens} dataset. We collect three datasets: one of the car with normal painting, one of the car with CNCA texture (We choose CNCA because it achieve the best attack among all the baselines in the simulation results in Table \ref{sim-attack-table1}.), and the one of the car with our BEVCA camouflage. We evaluate attack success rates across four victim models: BEVFormer-base, BEVFormer-small, BEVDet-R50, and BEVDet-R50-CBGS. BEVFormer \cite{li2022BEVFormer} and BEVDet \cite{huang2021bevdet} are representatives for backward and feedforward BEV-based perception models, respectively \cite{li2023fb}. As shown in Table \ref{phy-table1}, BEVCA achieves substantially higher ASR than CNCA across all victim models with average of 51.18\% ASR in black-box settings, demonstrating our method can pose a severe safety threat in real-world applications. Furthermore, Figure \ref{phy_exp_examples} shows that our camouflaged car can be successfully undetected from the 3D detector in different camera viewpoints while the other surrounding objects (cars, scooters, pedestrians) are well detected. In summary, the physical evaluation results demonstrate that our method is transferable to the real world.

\subsection{Ablation Studies}\label{ablation}

\begin{table*}[h]

\caption{The attack performance of different adversarial loss functions across various 3D detection and BEV segmentation models.}
\label{ab-attack-table3}

\begin{center}
\begin{small}
\begin{sc}
\resizebox{1.0\textwidth}{!}{
\begin{tabular}{lcccccccc}
\toprule
\multirow{2}{*}{\textbf{Methods}} & \multicolumn{5}{c}{3D Detection } & \multicolumn{3}{c}{BEV Segmentation}  \\
\cmidrule(lr){2-6} \cmidrule(lr){7-9} 
                         & Base & Small & Tiny & Base-Seg-Det & Small-Seg-Det & LSS & CVT & SinBEVT  \\
\midrule
\( L_{det} \) &  0.564 &  0.683 &  0.498 &  0.728 &  0.597 &  0.450&  0.505& 0.467\\
\( L_{enc} \) &  0.392&  0.564&  0.428&  0.528&  0.529&  0.380&  0.371& 0.333\\
\( L_{head} \) &  0.471&  0.603&  0.485&  0.601&  0.529&  0.367&  0.426& 0.347\\
\( L_{bev} \) &  0.612 &  0.704 &  0.508 &  0.735 &  0.585 &  0.498&  0.588& 0.574\\
\( L_{bev} + \mathcal{M}_{bev} \) &  \textbf{0.191} &  \textbf{0.285} & \textbf{0.381} &  \textbf{0.329} & \textbf{0.329} & \textbf{0.253}& \textbf{0.307}& \textbf{0.258}\\
\bottomrule
\end{tabular}
}
\end{sc}
\end{small}
\end{center}

\end{table*}

\textbf{Effectiveness of different adversarial losses.} To validate the necessity of both (i) BEV-feature-level optimization and (ii) our masking scheme, we compare our approach (\(L_{bev} + \mathcal{M}_{bev}\)) with several alternative adversarial losses against the same target model (BEVFormer-base): \(L_{det}\) based on 3D detection scores, \(L_{enc}\) based on image-encoded features within the BEV encoder, \(L_{head}\) based on hidden-state features from the 3D detection head, \(L_{bev}\) based on the whole BEV features without applying masks.

Table \ref{ab-attack-table3} presents the attack performance across different models in both 3D object detection and BEV segmentation tasks. The results reveal several critical insights:  first, directly attacking 3D detection outputs (\(L_{det}\)) shows limited effectiveness and transferability, achieving similar attack performance to baselines like FCA; second, attacking non-BEV intermediate features (\(L_{enc}\) and \(L_{head}\)) yields substantially weaker attack performance compared to our method, indicating suboptimal attacks; most importantly, attacking BEV features without masking (\(L_{bev}\)) shows almost no attack effect, demonstrating that naive BEV-level attacks fail due to overwhelming noise from car-irrelevant regions. Since the target vehicle occupies only approximately 3\% of the entire BEV perception area (as shown in Figure \ref{fig:bev_diff_vs_gt}), gradients are dominated by noise from irrelevant regions when the entire BEV feature is used for loss computation. In contrast, our proposed method (\(L_{bev} + \mathcal{M}_{bev}\) ) achieves the strongest attack performance, which demonstrates superior transferability across all black-box models in both 3D detection and BEV segmentation tasks. These results empirically validate that: (1) attacking BEV features achieves both effective and transferable attack performance, and (2) our masking technique is essential for locating the car-texture-sensitive BEV regions, enabling effective adversarial optimization. 

In addition, we conduct further ablation analyses in the appendix to provide a more complete understanding of BEVCA. Specifically, we study sensitivity to key optimization hyperparameters (the TV weight \(\beta\) and BEV-mask parameters \(\Delta r,\Delta \theta\)) in Figure~\ref{fig:hp_beta_tuning}, and evaluate robustness under simple defense transformations in Table~\ref{tab:defense_eval}. These additional studies consistently support the conclusions of our main ablation.

\section{Conclusion}

We have proposed BEVCA, a novel adversarial camouflage attack framework targeting multi-view BEV-based 3D perception. Our framework can generate effective and transferable adversarial camouflage across various multi-view BEV-based 3D perception models and tasks. Our approach consists of a novel BEV-feature-based adversarial loss that enables transferable attacks, and a multi-view neural rendering module that facilitates effective gradient-based camouflage optimization against multi-view perception models. Extensive experiments on both 3D object detection and BEV segmentation tasks demonstrate that BEVCA consistently outperforms existing baselines against diverse black-box models under multi-view perception settings. Moreover, we validate the practical effectiveness of our approach through physical experiments on a full-size vehicle, highlighting its robustness in the real world.

\section{Limitations and Future Work}
\label{sec:limitations_future_work}

While our physical experiments validate BEVCA's real-world effectiveness, several challenges still remain. First, physical deployment introduces a sim-to-real domain gap caused by factors like UV map deformation, and lighting conditions. Systematically quantifying these variables is left for future work. Second, adversarial textures are inherently bound to a specific vehicle's 3D geometry; transferring the attack to a different vehicle class requires a one-time offline re-optimization. Finally, as autonomous driving evolves toward Vision-Language-Action (VLA) frameworks and World Models, BEV representations remain their one of their foundational spatial features. While our framework exposes immediate risks in BEV perception, investigating the robustness of VLA models against adversarial camouflage is a critical direction for future research.

\section{Impact Statement}
\label{sec:impact_statement}

This work is intended to advance the safety and robustness of autonomous driving (AD) systems. While the proposed adversarial attack method could potentially be misused by malicious users, it primarily serves as a valuable tool for strengthening AD systems through adversarial training (see Appendix Section~\ref{appendix:adversarial_training}), adversarial testing, and adversarial detection. By exposing critical vulnerabilities under realistic physical conditions, our approach ultimately contributes to improving the security and reliability of AD systems.

\section*{Acknowledgment}
\label{sec:acknowledgment}

This research is supported by the National Key R\&D Program of China (Grant Nos. 2026ZD1500404 and 2024YFE0215300), the National Natural Science Foundation of China (Grant Nos. 62376074 and 62674206), the Guangdong S\&T Program (Grant No. 2026B0101100003), and the Shenzhen Science and Technology Program (Grant Nos. KJZD20231023095959002, KCXFZ20230731093001002, KJZD20240903100505007, ZDCY20250901104705006, and KJZD20240903102901003). Yu Li and Daojing He are the corresponding authors of this article.

\bibliographystyle{acl_natbib}
\bibliography{neurips_2026}


\appendix

\section{Multi-View Camera Pose Transformation Details}
\label{appendix:multi_view_neural_rendering}

The key challenge for precise multi-view rendering of the target vehicle is to compute the transformation from each camera coordinate system to the target vehicle’s coordinate system. First, we can obtain the following geometric data from the CARLA API:

\begin{itemize}
    \item The global position and rotation of the ego vehicle: \( t_{ego}^{glob} \), \( R_{ego}^{glob} \).
    \item The global position and rotation of the target vehicle: \( t_{tar}^{glob} \), \( R_{tar}^{glob} \).
    \item The position and rotation of the ego camera in the ego vehicle frame: \( t_{cam}^{ego} \), \( R_{cam}^{ego} \).
\end{itemize}

Since CARLA employs a left-handed coordinate system while PyTorch3D uses a right-handed one, we first convert all relevant data accordingly through a proper handedness transformation.
\begin{equation}
\hat{t} = M \cdot t, \;
\hat{R} = M \cdot R \cdot M, \; 
\text{where } M = diag(1, -1, 1)
\end{equation}

where \(\hat{t}\) and \(\hat{R}\) denote transformed quantities in the right-handed coordinate system. With the above information, we can construct the following transformation matrices:
\begin{equation}
T_{tar \leftarrow glob} = \begin{bmatrix} \hat{R}_{tar}^{glob} & \hat{t}_{tar}^{glob} \\ 0_{1 \times 3} & 1 \end{bmatrix}^{\top}, \; T_{glob \leftarrow ego} = \begin{bmatrix} \hat{R}_{ego}^{glob} & \hat{t}_{ego}^{glob} \\ 0_{1 \times 3} & 1 \end{bmatrix}
\end{equation}

We then obtain the transformation matrix from the ego vehicle coordinate system to the target vehicle coordinate system as:
\begin{equation}
T_{tar \leftarrow ego} = T_{tar \leftarrow glob} \cdot T_{glob \leftarrow ego}
\end{equation}

To obtain the camera pose matrix in the ego vehicle coordinate system, we construct it as follows:
\begin{equation}
\Phi_{cam} =  
\begin{bmatrix} 
\hat{R}_{cam}^{ego}\cdot R_D & \hat{t}_{cam}^{ego} \\ 
0_{1 \times 3} & 1 
\end{bmatrix}, 
\quad
R_D = \begin{bmatrix}
0 & 0 & 1 \\
1 & 0 & 0 \\
0 & 1 & 0
\end{bmatrix}
\end{equation}
where \(R_{D}\) is a transformation matrix that converts Carla's camera convention (x-forward, z-up) to PyTorch3D's camera convention (z-forward, y-up). Eventually, we compute the camera pose \(\Phi_{r}\) in the target vehicle coordinate system, which is required as input by PyTorch3D, via:
\begin{equation}
\Phi_{r} = T_{tar \leftarrow ego} \cdot \Phi_{cam}
\end{equation}

For multi-view rendering, We apply the above camera parameter transformation to all six ego-mounted cameras and perform image rendering as defined in Eq ~\ref{1}.

\section{Physical World Experiments}

\begin{figure*}[t]
\centering
\includegraphics[width=0.6\textwidth]{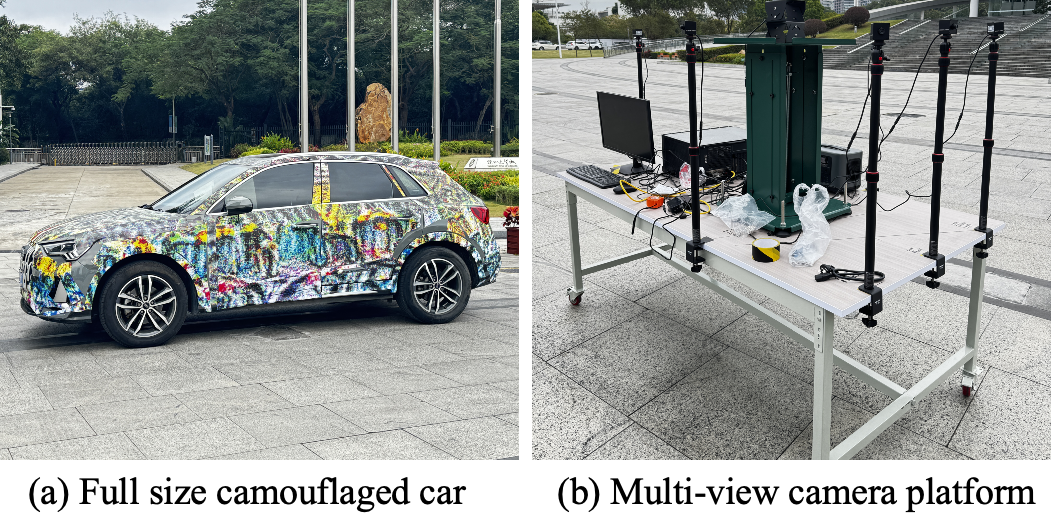}

\caption{The physical experiment setup: (a) full size camouflaged car; (b) multi-view camera data collection platform.
  }
\label{phy_exp_setup}
\end{figure*}

We have conducted the physical experiment to validate our method in the real world. Figure \ref{phy_exp_setup} shows the setup for our physical experiments. We print our generated BEVCA camouflage in a 2D UV map and stick it to a real-life-size Audi vehicle. We also built a simple multi-view camera data collection platform to collect real-life multi-view data. We set up six camera sensors on this platform, following the camera settings (positions and angles) from nuScenes. With this platform, we collect three datasets: one of the car with normal painting, one with CNCA texture, and one of the car with our BEVCA camouflage. Each dataset has 150 samples with different azimuths and distances towards the ego position.

\section{Multi-view Attack Visualization}

\begin{figure}[htbp]
\centering
\includegraphics[width=1.0\linewidth]{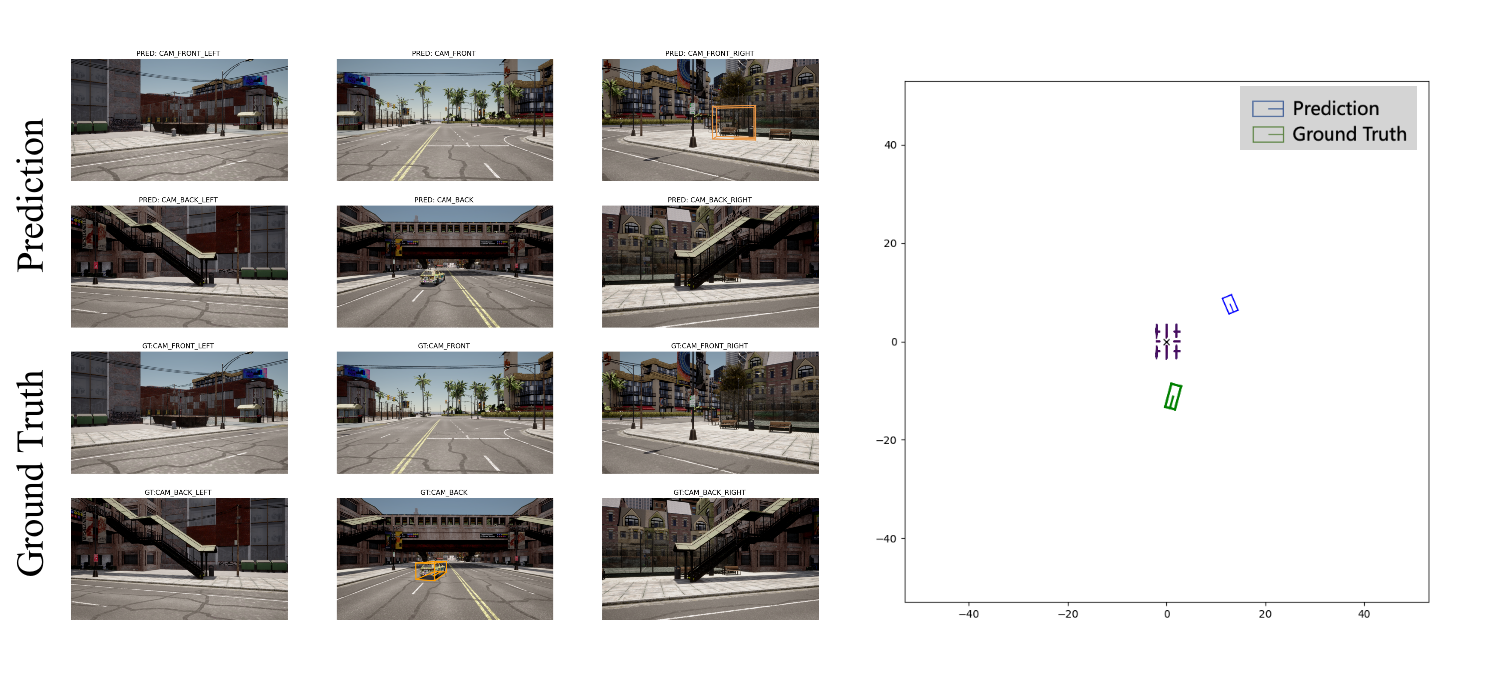}
\vspace{-3mm}
\includegraphics[width=1.0\linewidth]{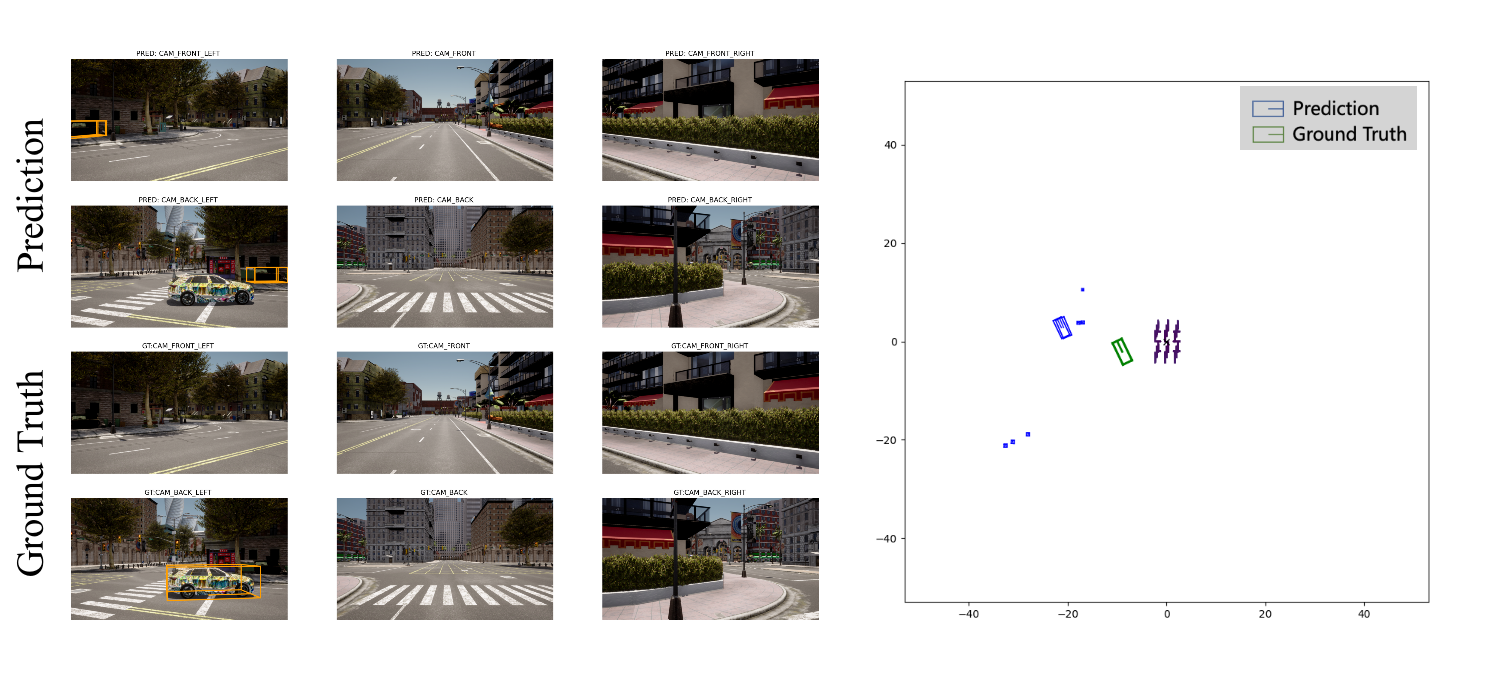}
\vspace{-3mm}
\includegraphics[width=1.0\linewidth]{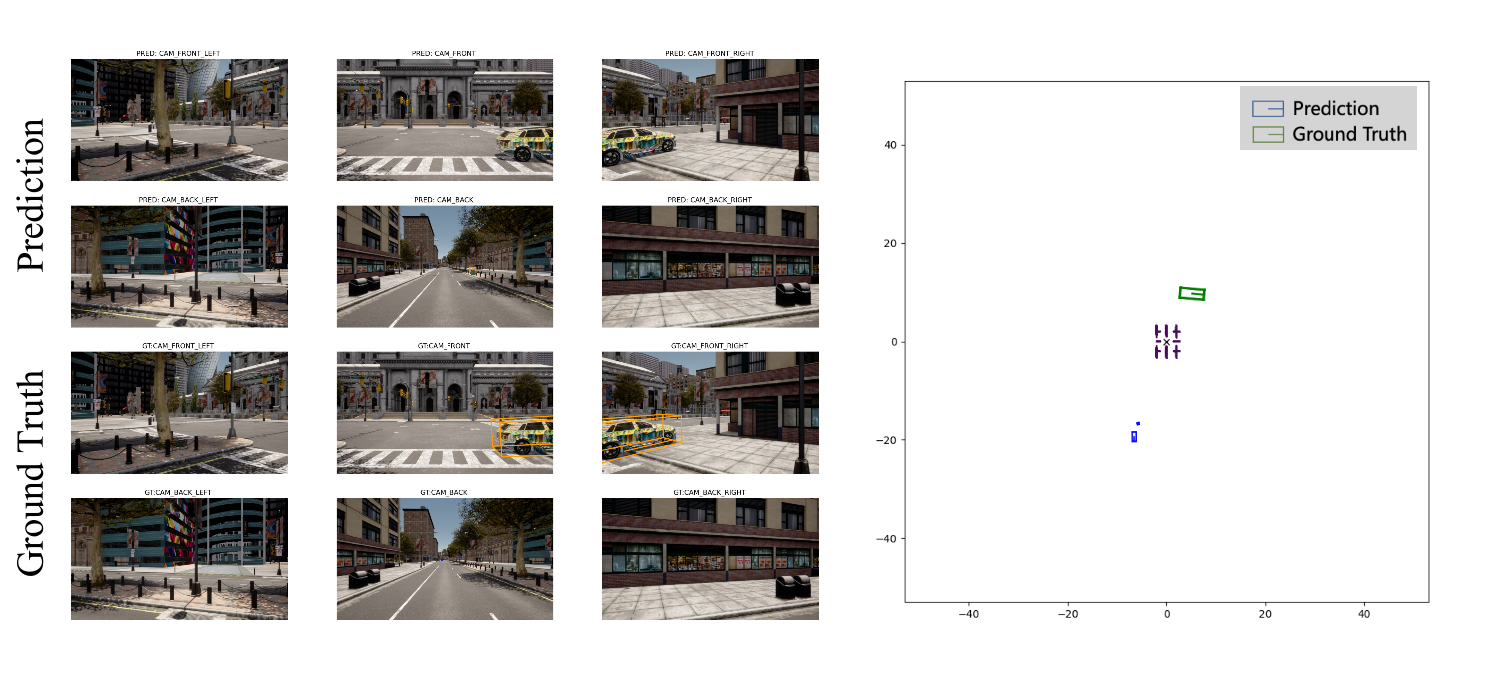}
\caption{Visualization results of our camouflage attack against 3D object detection tasks. We can observe that the camouflaged vehicle leads to inaccurate bounding boxes or complete “disappearance” from the detector compared to the ground truths.}
\label{fig:digital_attack_examples}
\end{figure}

\begin{figure}[htbp]
\centering
\includegraphics[width=1.0\linewidth]{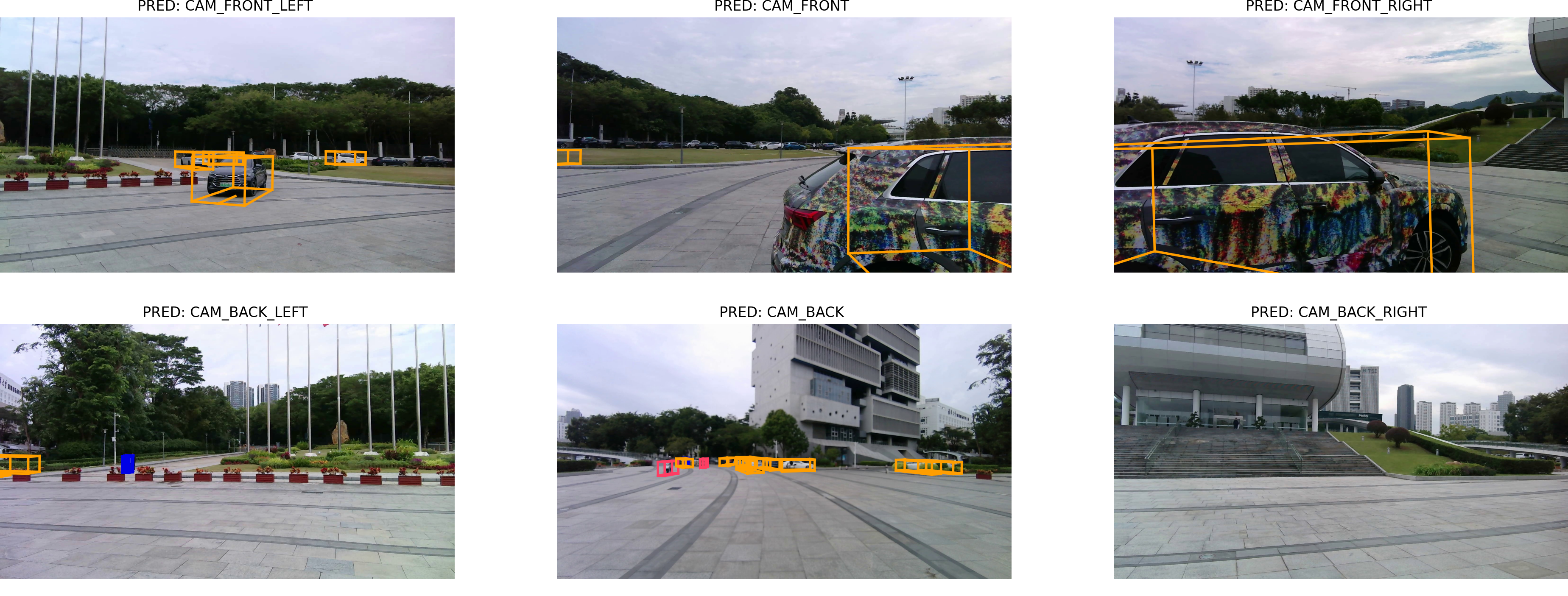}
\includegraphics[width=1.0\linewidth]{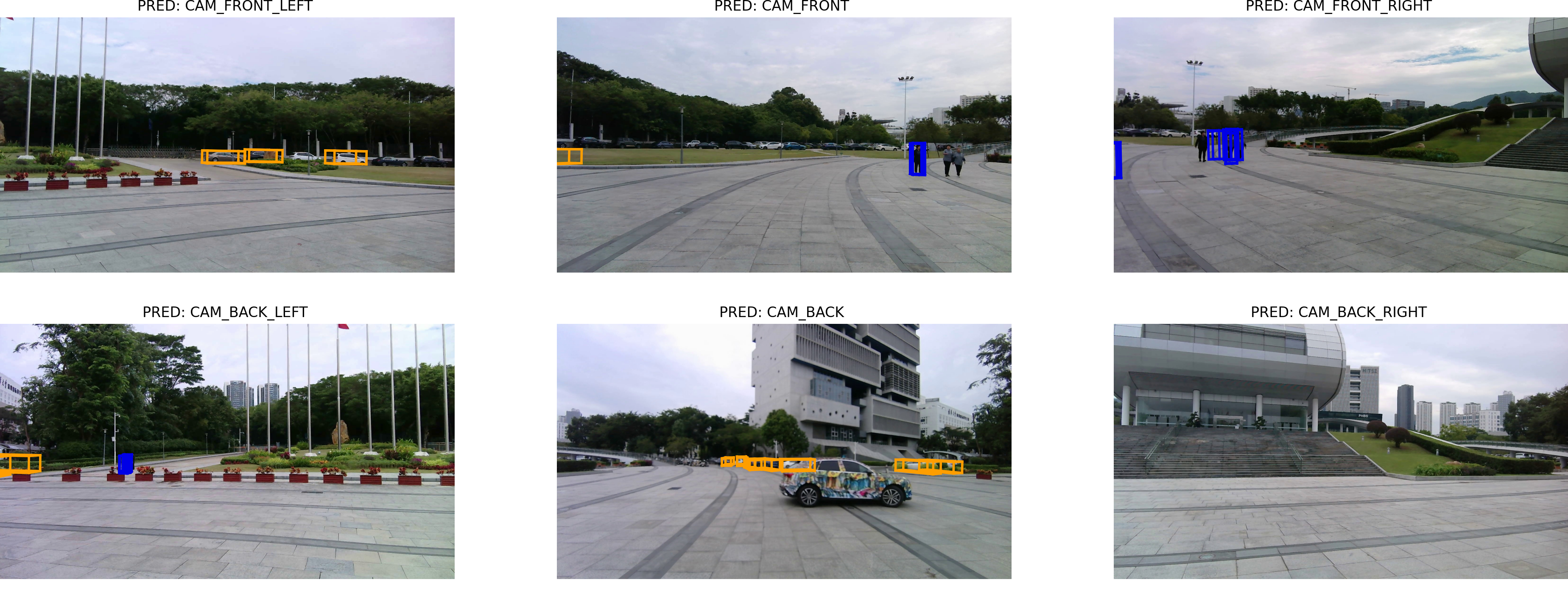}
\includegraphics[width=1.0\linewidth]{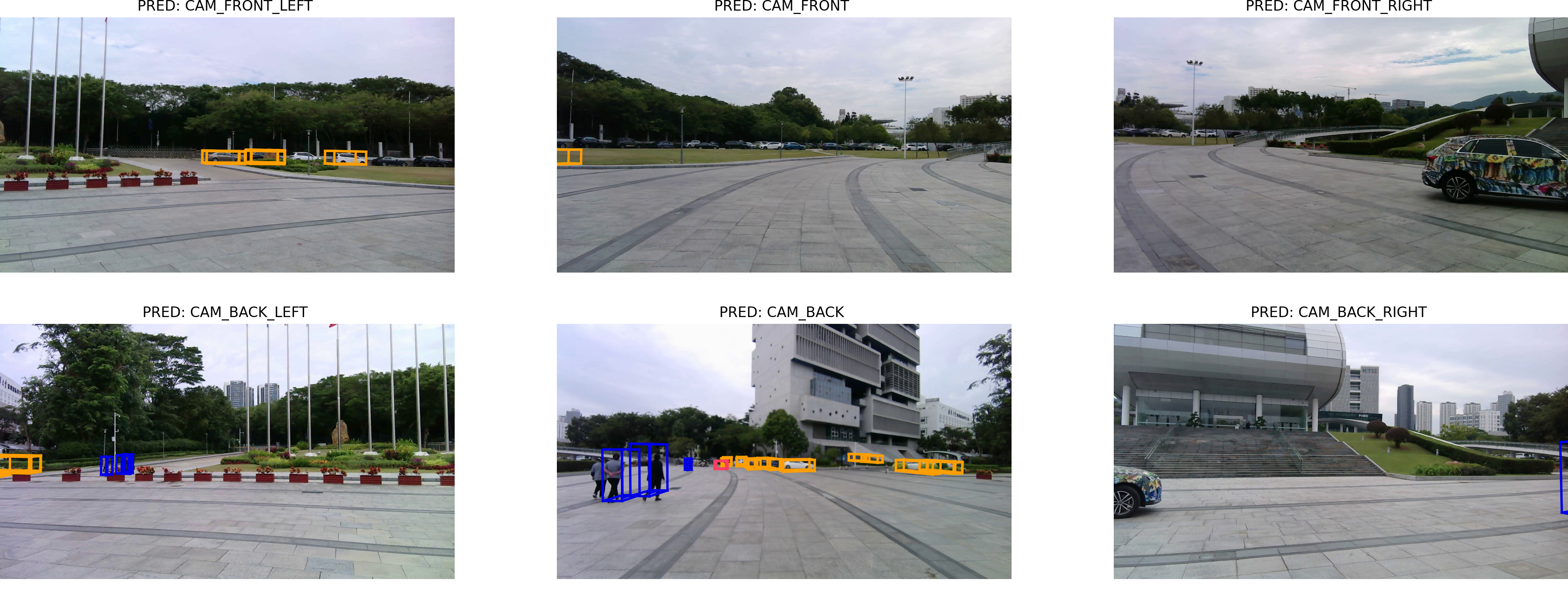}
\includegraphics[width=1.0\linewidth]{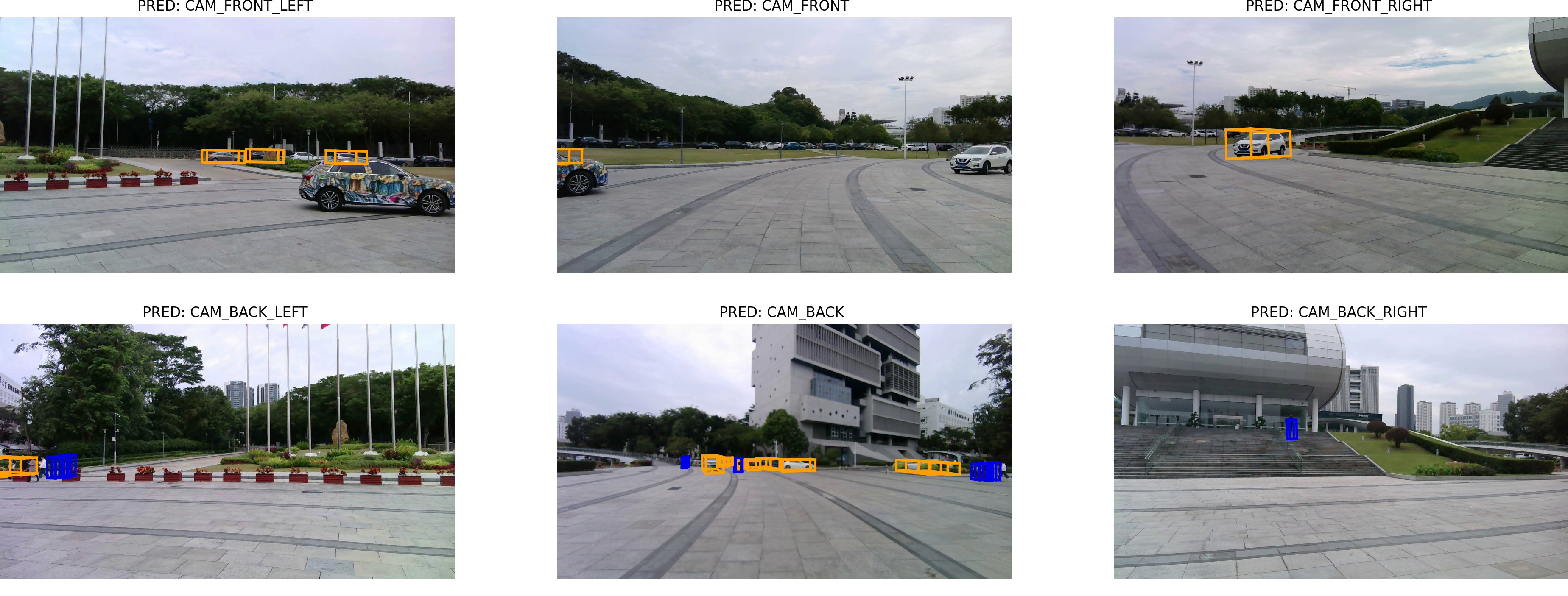}

\caption{Additional visualization results of our camouflage attack against 3D object detection tasks in the physical world.}
\label{fig:physical_attack_examples}
\end{figure}

As shown in Figure \ref{fig:digital_attack_examples}, we provide some examples of our camouflage attack against 3D object detection in the digital world. Compared with the ground truth label, we can see that the camouflaged vehicle leads to inaccurate bounding boxes or complete “disappearance” from the detector. Our camouflage can achieve successful attacks when the target vehicle appears in not only one but also across views. As shown in Figure \ref{fig:physical_attack_examples}, we provide  examples of our camouflage attack against 3D object detection with full camera views in the real world evaluation.
 
\section{Enhancing Model Robustness via Adversarial Training}
\label{appendix:adversarial_training}
Beyond identifying model vulnerabilities, our attack also provides a practical mechanism for enhancing the robustness of BEV-based perception models through adversarial training. To demonstrate this, we construct an adversarial dataset consisting of 500 samples generated using our camouflage. To avoid overlap with the evaluation data, the dataset is created under camera configurations different from those used in the test set. We then fine-tune the original BEV perception models on this adversarial dataset. The resulting performance improvements are summarized in Table~\ref {tab:adversarial_training}.

The results clearly demonstrate that our adversarial camouflage severely degrades the performance of original BEV-based perception models, exposing critical robustness weaknesses. For instance, BEVFormer-base’s detection score drops from 0.618 to 0.191, while CVT falls from 0.609 to 0.307 under attack. Importantly, adversarial training with our adversarial examples substantially restores performance across all models, specifically raising BEVFormer-base to 0.615 and CVT to 0.567, confirming that our method serves not only as a strong attack but also as an effective tool for enhancing model robustness.

\begin{table*}[t]
\caption{Performance comparison of BEV-based perception models under normal painting (without attack), BEVCA attack, and after adversarial training. The metrics for 3D detection and BEV segmentation are AP and IoU of the vehicle class, respectively.}
\label{tab:adversarial_training}

\begin{center}
\begin{small}
\begin{sc}
\resizebox{\textwidth}{!}{ 
\begin{tabular}{lcccccccc}
\toprule
\multirow{2}{*}{\textbf{Methods}} & \multicolumn{5}{c}{3D Detection} & \multicolumn{3}{c}{BEV Segmentation } \\
\cmidrule(lr){2-6} \cmidrule(lr){7-9}
                         & Base & Small & Tiny & Base-Seg-Det & Small-Seg-Det & LSS & CVT & SinBEVT \\
\midrule
Normal                   & 0.618& 0.721& 0.514& 0.750& 0.634& 0.509& 0.609& 0.570\\
BEVCA                      & 0.191& 0.285& 0.381& 0.329& 0.331& 0.253& 0.307&0.273\\
AdvTrain                      & 0.615& 0.479& 0.416& 0.683& 0.599& 0.614& 0.567&0.549\\
\bottomrule
\end{tabular}
}
\end{sc}
\end{small}
\end{center}

\end{table*}

\section{Multi-view robust attack.} 

\begin{table}[t]
\caption{The attack performance for multi-azimuth and multi-distance. Notice the AP of the normal painted car is 0.618.}
\label{multi-robust-table2}
\begin{center}
\begin{small}
\begin{sc}
\resizebox{0.6\columnwidth}{!}{
\begin{tabular}{cccccc}
\toprule
\multirow{2}*{\textbf{Azimuth ($^\circ$)}} & \multicolumn{4}{c}{\textbf{Distance}} & \multirow{2}*{\textbf{Avg}} \\
\cmidrule(lr){2-5}
& 10 & 15 & 25 & 40 & \\
\midrule
        0 & 0.278& 0.129& 0.468& 0.376& 0.313\\
        60 & 0.005& 0.184& 0.311& 0.278& 0.195\\
        120 & 0.206& 0.332& 0.167& 0.218& 0.231\\
        180 & 0.123& 0.641& 0.404& 0.001& 0.292\\
        240 & 0.242& 0.293& 0.300& 0.195& 0.258\\
        300 & 0.088& 0.282& 0.347& 0.197& 0.228\\
        \midrule
        \textbf{Avg} & 0.157& 0.310& 0.333& 0.211& 0.253\\
\bottomrule
\end{tabular}
}
\end{sc}
\end{small}
\end{center}
\end{table}

To evaluate the robustness of our adversarial camouflage under varying target vehicle positions, we conduct comprehensive experiments across different distances and azimuths relative to the ego vehicle. Specifically, we generate a dataset where the target vehicle appears at distances of [10, 15, 25, 40] meters, aligned with the BEVFormer model’s perception range of up to 51.2 meters. For azimuths, we sample [0°, 60°, 120°, 180°, 240°, 300°]. For each distance-azimuth combination, we create 40 samples with varying background scenes across different CARLA towns to ensure diversity. We use the BEVFormer base model for evaluation, as other detectors exhibit similar trends.

The results are shown in Table~\ref{multi-robust-table2}. The total mean Car AP value is comparable to the value reported in Table~\ref{sim-attack-table1}.  We also observe that with the increase of the distance, the AP first prone to increase at a distance of 10, after that the AP prone to decrease after a distance of 25. This trend can be attributed to two factors: (1) with the increase of the distance, fine-grained adversarial camouflage patterns are difficult for the detection model to identify, causing the decrease of the attack performance; and (2) the detection performance of the model declines at long ranges, causing the AP to decrease again. Nevertheless, our camouflage maintains strong attack effectiveness across all tested configurations, demonstrating its robustness in multi-view settings.

\section{Effectiveness of hyperparameters during optimization.} 



\begin{figure}[t]
\centering
\includegraphics[width=\textwidth]{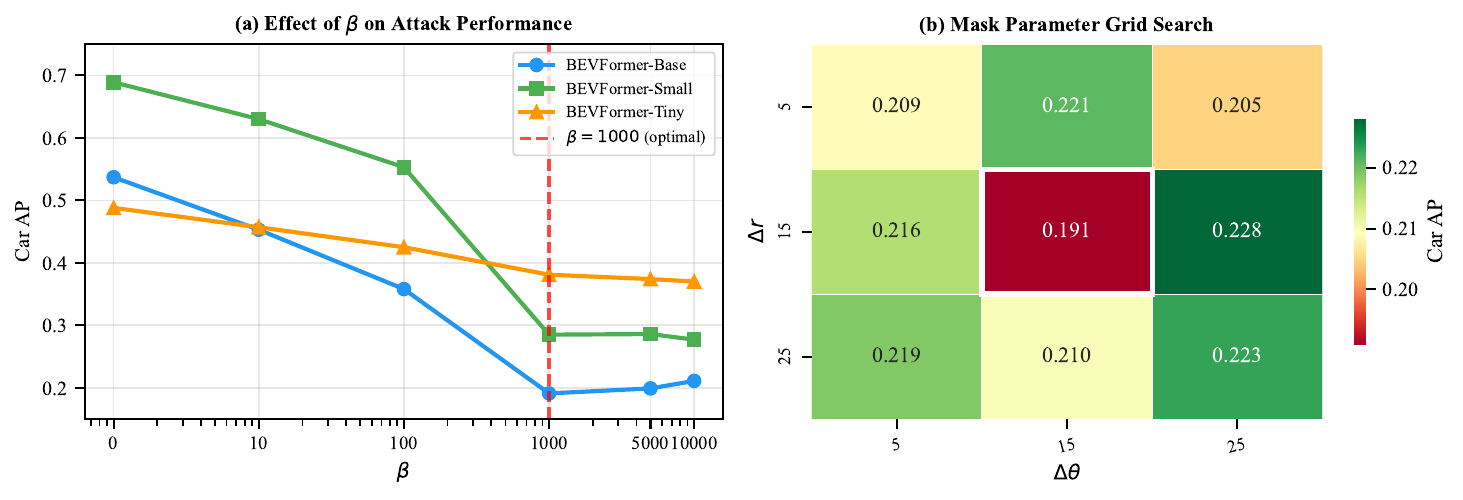}
\caption{Hyperparameter analysis: (a) Effect of $\beta$ on attack performance; (b) Grid search over mask parameters $\Delta r$ and $\Delta \theta$.}
\label{fig:hp_beta_tuning}
\end{figure}

We investigate the impact of the main hyperparameters during camouflage optimization related to total variation loss and BEV mask. Specifically, we vary the hyperparameter \(\beta\) that controls the strength of the total variation constraint, and report the results in figure~\ref{fig:hp_beta_tuning} (a) with BEVFormer series models. The results show that incorporating the total variation loss consistently improves attack performance. When 
\(\beta\) is too small, the generated textures lack coherent structure and fail to deliver strong adversarial signals—especially at long ranges. As \(\beta\) increases, the attack performance improves steadily, highlighting the importance of enforcing total variation in the camouflage pattern. For BEV mask settings, we run hyperparameter tuning experiments with the BEVFormer-base model on two tunable thresholds \(\Delta r\) and \(\Delta \theta\)  to find suitable values such that it can locate the relevant area accurately. The results from Figure~\ref{fig:hp_beta_tuning} (b) show that \(\Delta r\) with value of 15 and \(\Delta \theta\) with values of 15 give the best results. We fix these settings throughout our experiments.

\section{Robustness Against Simple Defenses.} 

To contextualize the practical risk of our attack framework, we evaluate BEVCA against three standard, lightweight defensive mechanisms applied to the BEVFormer-base model: BEV Feature Smoothing (Gaussian), Input Camera Image Smoothing (Gaussian), and JPEG Compression. 

As shown in Table~\ref{tab:defense_eval}, while aggressive defenses (e.g., severe JPEG compression or heavy smoothing) can marginally degrade the attack's effectiveness, they concurrently introduce non-trivial performance trade-offs on clean data (Normal mAP). BEVCA consistently maintains a massive performance gap over the baseline across all tested defensive parameters, demonstrating robust evasion capabilities that cannot be mitigated by simple input or feature transformations.

\begin{table}[h]
\centering
\caption{Performance of BEVFormer-base under BEVCA attack when employing various simple defense mechanisms. The Performance Gap indicates the difference between Normal mAP and BEVCA mAP.}
\label{tab:defense_eval}
\resizebox{\columnwidth}{!}{%
\begin{tabular}{llccc}
\toprule
\textbf{Defense Method} & \textbf{Parameter} & \textbf{Normal mAP} & \textbf{BEVCA mAP} & \textbf{Performance Gap} \\ \midrule
No Defense & - & 0.618 & 0.191 & 0.427 \\ \midrule
\multirow{3}{*}{BEV Feature Smoothing} 
 & $\sigma=0.5$ & 0.616 & 0.189 & 0.427 \\
 & $\sigma=1.0$ & 0.594 & 0.175 & 0.419 \\
 & $\sigma=2.0$ & 0.482 & 0.121 & 0.361 \\ \midrule
\multirow{3}{*}{Input Smoothing} 
 & $\sigma=0.5$ & 0.617 & 0.171 & 0.446 \\
 & $\sigma=1.0$ & 0.614 & 0.165 & 0.449 \\
 & $\sigma=2.0$ & 0.612 & 0.283 & 0.329 \\ \midrule
\multirow{3}{*}{JPEG Compression} 
 & $QF=75$ & 0.615 & 0.150 & 0.465 \\
 & $QF=50$ & 0.620 & 0.136 & 0.484 \\
 & $QF=25$ & 0.620 & 0.113 & 0.507 \\ \bottomrule
\end{tabular}%
}
\end{table}

\section{Attack Performance on Original Targeted Models}

\begin{table*}[t]
\caption{Comparison of the effectiveness of attacks across various original 3D detection and BEV segmentation models. The metrics for 3D detection and BEV segmentation are AP and IoU of the vehicle class, respectively.}
\label{sim-attack-original-table}

\begin{center}
\begin{small}
\begin{sc}
\resizebox{\textwidth}{!}{ 
\begin{tabular}{lcccccccc}
\toprule
\multirow{2}{*}{\textbf{Methods}} & \multicolumn{5}{c}{3D Detection} & \multicolumn{3}{c}{BEV Segmentation } \\
\cmidrule(lr){2-6} \cmidrule(lr){7-9}
                         & Base & Small & Tiny & Base-Seg-Det & Small-Seg-Det & LSS & CVT & SinBEVT \\
\midrule
Normal                   & 0.309& 0.324& 0.243& 0.321& 0.333& 0.209& 0.234& 0.238\\
FCA                      & 0.306& 0.336& 0.263& 0.32& 0.337& 0.204& 0.200&0.208\\
DTA                      & 0.260& 0.296& 0.249& 0.294& 0.318& 0.175& 0.168&0.195\\
ACTIVE                   & 0.236& 0.264& 0.220& 0.265& 0.292& 0.178& 0.183&0.195\\
RAUCA                    & 0.259& 0.29& 0.204& 0.295& 0.295& 0.162& 0.157&0.164\\
CNCA                     & 0.227& 0.253& 0.185& 0.245& 0.284& 0.157& 0.201&0.181\\
AdvPatch                 & 0.307& 0.310& 0.235& 0.321& 0.331& 0.193& 0.213&0.202\\
\midrule
BEVCA                    & \textbf{0.080}& \textbf{0.126}& \textbf{0.158}& \textbf{0.148}& \textbf{0.171}& \textbf{0.123}& \textbf{0.065}& \textbf{0.106}\\
\bottomrule
\end{tabular}
}
\end{sc}
\end{small}
\end{center}

\end{table*}

Table ~\ref{sim-attack-original-table} shows the comparison of the effectiveness of different attacks across different 3D detection and BEV segmentation original models, which are trained on the nuScenes dataset \cite{Caesar2020NuScens}. Compared to Table ~\ref{sim-attack-table1}, we can notice the performance gap for each attack-model pair between these two tables due to the domain gap between simulation and reality. However, BEVCA still outperforms all the baselines, proving the effectiveness and transferability of our method.

\section{Usage of LLMs}
\label{sec:usage_of_llms}
We have mainly used LLMs to help us with the editing of this paper and write code to anlayze and visulize the experiment resutls. However, the core ideas and methodology of this paper are all original and not influenced by LLMs.

\newpage
\section*{NeurIPS Paper Checklist}

\begin{enumerate}

\item {\bf Claims}
    \item[] Question: Do the main claims made in the abstract and introduction accurately reflect the paper's contributions and scope?
    \item[] Answer: \answerYes{} 
    \item[] Justification: The abstract and introduction explicitly state our core contributions and scope of our paper.

\item {\bf Limitations}
    \item[] Question: Does the paper discuss the limitations of the work performed by the authors?
    \item[] Answer: \answerYes{} 
    \item[] Justification: We discuss the limitations in a dedicated section \ref{sec:limitations_future_work} .

\item {\bf Theory assumptions and proofs}
    \item[] Question: For each theoretical result, does the paper provide the full set of assumptions and a complete (and correct) proof?
    \item[] Answer: \answerNA{} 
    \item[] Justification: Our work result are mostly experimental and does not involve obtaining theoretical result.

    \item {\bf Experimental result reproducibility}
    \item[] Question: Does the paper fully disclose all the information needed to reproduce the main experimental results of the paper to the extent that it affects the main claims and/or conclusions of the paper (regardless of whether the code and data are provided or not)?
    \item[] Answer: \answerYes{} 
    \item[] Justification: We have open-sourced our code and described the detailed information to reproduce our experiment in the experiment setting section \ref{sec:experiment_setting}.

\item {\bf Open access to data and code}
    \item[] Question: Does the paper provide open access to the data and code, with sufficient instructions to faithfully reproduce the main experimental results, as described in supplemental material?
    \item[] Answer: \answerYes{} 
    \item[] Justification: We have provided the link for the related code and data in the abstract for reproducing our results.

\item {\bf Experimental setting/details}
    \item[] Question: Does the paper specify all the training and test details (e.g., data splits, hyperparameters, how they were chosen, type of optimizer) necessary to understand the results?
    \item[] Answer: \answerYes{} 
    \item[] Justification: We have described our experiment setting and details in the section \ref{sec:experiment_setting}.

\item {\bf Experiment statistical significance}
    \item[] Question: Does the paper report error bars suitably and correctly defined or other appropriate information about the statistical significance of the experiments?
    \item[] Answer: \answerYes{} 
    \item[] Justification: We report the mAP, IoU and ASR for the attack performance in the table \ref{sim-attack-table1} and \ref{phy-table1}, which are calculate by dataset s with size at least over than 150. Therefore the reported results have statistical significance. 

\item {\bf Experiments compute resources}
    \item[] Question: For each experiment, does the paper provide sufficient information on the computer resources (type of compute workers, memory, time of execution) needed to reproduce the experiments?
    \item[] Answer: \answerYes{} 
    \item[] Justification: We have described the compute resources needed for our experiment in the section \ref{sec:experiment_setting}.
    
\item {\bf Code of ethics}
    \item[] Question: Does the research conducted in the paper conform, in every respect, with the NeurIPS Code of Ethics \url{https://neurips.cc/public/EthicsGuidelines}?
    \item[] Answer: \answerYes{} 
    \item[] Justification: We have reviewd the NeurIPS code of Ethics and confirm our research are conducted aligned with the NeurIPS code of Ethics.

\item {\bf Broader impacts}
    \item[] Question: Does the paper discuss both potential positive societal impacts and negative societal impacts of the work performed?
    \item[] Answer: \answerYes{} 
    \item[] Justification: We have discussed both the potential positive and negative societal impact in the section \ref{sec:impact_statement}.
    
\item {\bf Safeguards}
    \item[] Question: Does the paper describe safeguards that have been put in place for responsible release of data or models that have a high risk for misuse (e.g., pre-trained language models, image generators, or scraped datasets)?
    \item[] Answer: \answerYes{} 
    \item[] Justification: We have discussed safeguards to enhance the AI systems against our proposed attack in the section \ref{sec:impact_statement}.

\item {\bf Licenses for existing assets}
    \item[] Question: Are the creators or original owners of assets (e.g., code, data, models), used in the paper, properly credited and are the license and terms of use explicitly mentioned and properly respected?
    \item[] Answer: \answerYes{} 
    \item[] Justification: We confirmed that all the assets that we used in our worked are properly credited. The related license and terms of use explicitly are mentioned and properly respected.

\item {\bf New assets}
    \item[] Question: Are new assets introduced in the paper well documented and is the documentation provided alongside the assets?
    \item[] Answer: \answerYes{} 
    \item[] Justification: The new assets introduced in this paper is multi-view neural rendering module and BEV-feature-based adversarial loss. We have provided the detailed documentation for the module and loss function in the section \ref{appendix:multi_view_neural_rendering} and \ref{bev-feature-attack}.

\item {\bf Crowdsourcing and research with human subjects}
    \item[] Question: For crowdsourcing experiments and research with human subjects, does the paper include the full text of instructions given to participants and screenshots, if applicable, as well as details about compensation (if any)? 
    \item[] Answer: \answerNo{} 
    \item[] Justification: We do not involve crowdsourcing and research with human subjects in this paper.

\item {\bf Institutional review board (IRB) approvals or equivalent for research with human subjects}
    \item[] Question: Does the paper describe potential risks incurred by study participants, whether such risks were disclosed to the subjects, and whether Institutional Review Board (IRB) approvals (or an equivalent approval/review based on the requirements of your country or institution) were obtained?
    \item[] Answer: \answerNo{} 
    \item[] Justification: We do not involve research with human subjects in this paper.

\item {\bf Declaration of LLM usage}
    \item[] Question: Does the paper describe the usage of LLMs if it is an important, original, or non-standard component of the core methods in this research? Note that if the LLM is used only for writing, editing, or formatting purposes and does \emph{not} impact the core methodology, scientific rigor, or originality of the research, declaration is not required.
    \item[] Answer: \answerYes{} 
    \item[] Justification: We discuss the usage of LLMs in the section \ref{sec:usage_of_llms}.

\end{enumerate}

\end{document}